\documentclass[runningheads]{llncs}
\usepackage[latin9]{inputenc}
\usepackage{xcolor}
\usepackage{amsmath}
\usepackage{amssymb}
\usepackage{graphicx}

\makeatletter

\providecommand{\tabularnewline}{\\}

\usepackage{tikz}
\usepackage{comment}
\usepackage{color}
\usepackage{microtype}

\usepackage[accsupp]{axessibility}

\mainmatter

\author{Hoang-Anh Pham$^{1}$, Thao Minh Le$^{1}$, Vuong Le$^{1}$, Tu Minh Phuong$^{2}$, Truyen Tran$^{1}$}
\institute{$^{1}$Applied Artificial Intelligence Institute, Deakin University, Australia \\
$^{2}$Posts and Telecommunications Institute of Technology, Vietnam\\
\tt\small $^{1}$\{phamhoan, thao.le, vuong.le, truyen.tran\}@deakin.edu.au\\ $^{2}$phuongtm@ptit.edu.vn}
 
\makeatother

\begin{document}
\global\long\def\ModelName{\mathtt{COST}}%
\global\long\def\ReasonUnit{\mathtt{R3}}%

\title{Video Dialog as Conversation about Objects Living in Space-Time}
\maketitle
\begin{abstract}
It would be a technological feat to be able to create a system that
can hold a meaningful conversation with humans about what they watch.
A setup toward that goal is presented as a video dialog task, where
the system is asked to generate natural utterances in response to
a question in an ongoing dialog. The task poses great visual, linguistic,
and reasoning challenges that cannot be easily overcome without an
appropriate representation scheme over video and dialog that supports
high-level reasoning. To tackle these challenges we present a new
object-centric framework for video dialog that supports neural reasoning
dubbed $\ModelName$--which stands for \emph{Conversation about Objects
in Space-Time}. Here dynamic space-time visual content in videos is
first parsed into object trajectories. Given this video abstraction,
$\ModelName$ maintains and tracks object-associated \emph{dialog
states}, which are updated upon receiving new questions. Object interactions
are dynamically and conditionally inferred for each question, and
these serve as the basis for relational reasoning among them. $\ModelName$
also maintains a history of previous answers, and this allows retrieval
of relevant object-centric information to enrich the answer forming
process. Language production then proceeds in a step-wise manner,
taking the context of the current utterance, the existing dialog,
and the current question. We evaluate $\ModelName$ on the AVSD test
splits (DSTC7 and DSTC8), demonstrating its competitiveness against
state-of-the-arts.

\end{abstract}

\section{Introduction}

\begin{figure}[t]
\begin{centering}
\includegraphics[width=0.9\textwidth]{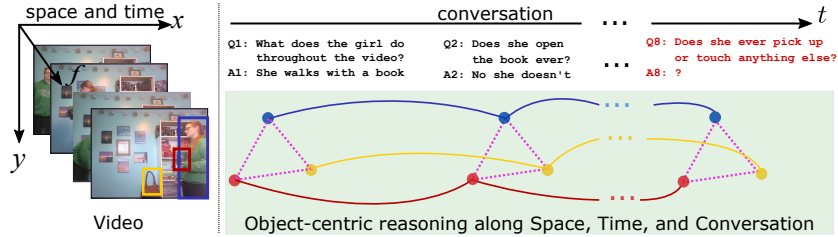}
\par\end{centering}
\centering{}\caption{We introduce $\protect\ModelName$, an object-centric reasoning framework
that collects clues along with the Space and Time dimensions of the
Video and the Conversation dimension of the dialog toward reliable
QA. \label{fig:teaser}}
\end{figure}

It is a hallmark of visual intelligence to build a system that can
hold a meaningful conversation with humans about a video. A system
of this capability would be a strong contender for passing the visual
Turing test \cite{geman2015visual}. Posed as video dialog \cite{alamri2019audio,geng2021dynamic},
this task challenges the current arts due to its sheer complexity
on many fronts. A dialog has its natural flow through multiple turns,
each of which builds upon the previous questions and answers. This
demands deep linguistic understanding about, and keeping track of,
what has been said and grounded on the visual concepts found in the
video, and then analyzing the new question in this newly established
context. Given the question semantics and its constituent words, generating
a linguistic answer necessitates symbolic grounding and visual reasoning
through the complex space-time structure of the video, where in multiple
objects interact in a dynamic manner.

Video dialog is inherently harder than the task of visual dialog over
static images \cite{Das2019} due to the temporal dynamics of the
scene \cite{alamri2019audio}. It is also harder than the standard
setting of video QA \cite{le2020hierarchical} since the next question
may not be comprehensible without maintaining a history of, and referring
to the previous answers. There have been several attempts to tackle
these challenges. Early attempts \cite{hori2019end,le2019end,nguyen2018film,sanabria2019cmu}
encode the dialog flow using recurrent neural networks. Later methods
\cite{le2020multimodal,lee2020dstc8} resort to Transformers for better
distant dependencies as well as cross-modality relations. More recent
methods make use of graphs as a representation of dialog structures
\cite{geng2021dynamic,le2021learning} and co-reference \cite{kim2021structured}
which achieved the new state-of-the-art result for this task. However,
we have only scratched the surface of what is possible and the principal
challenges remain.

A plausible pathway toward solving the remaining video dialog challenges
is through high-level, object-centric representation and reasoning
as seen in human visual cognition: Humans see objects and agency as
the core ``living'' constructs with natural compositionally, permanency,
temporal dynamics, and $n$-body interactions \cite{spelke2007core}.
Importantly, the object-centric approach has recently been found to
be essential for reasoning in Visual QA \cite{le2020dynamic} and
Video QA \cite{dang2021hierarchical}, thanks to the ease of binding
linguistic concepts to visual regions.

To this end, we propose a new object-centric framework dubbed $\ModelName$
\sloppy(\textbf{C}onversation about \textbf{O}bjects in \textbf{S}pace-\textbf{T}ime)
for video dialog. Video is first parsed into a set of object trajectories
throughout the video across the spatial dimensions of the frame and
the temporal spans of object lives. Central to $\ModelName$ is a
model of the object states dynamics as the conversation progresses\footnote{This is related to, but distinct from, the dialog state tracking in
typical task-oriented dialogs in NLP \cite{gao2019dialog}}. In particular, $\ModelName$ maintains a \emph{recurrent system
of} \emph{dialog-induced object states} throughout the course of conversation.
For each new question, the object lives will be consulted through
semantic word-object grounding and selective frame attention, producing
question-guided object representations. These serve as input for the
\emph{dialog state recurrent networks} to generate updated dialog
states. These states are used to construct a \emph{question-specific
interaction matrix} among objects, enabling relational reasoning among
them. The results are coupled with the answers from the previous conversational
turns to produce new representations, which are then decoded into
the response utterance. See Fig\@.~\ref{fig:teaser} for an illustration
of $\ModelName$ in action.

We evaluate our proposed $\ModelName$ on the public AVSD dataset
with two different test splits at DSTC7 and DSTC8. Experimental results
show that $\ModelName$ is highly competitive against rivals.

\section{Related Work}

\paragraph{Visual and video dialog}

The visual dialog task and an accompanied dataset (VisDial) were first
introduced in \cite{Das2019}. The task requires both conversational
and visual reasoning ability over multiple turns . Like the precursor
Visual QA task, visual dialog needs deep understanding of the visual
concepts and relations in the images, and reasoning about them in
response to the current question. A unique challenge in visual dialog
is the co-referrence problem of linguistic information between dialog
turns. The early works tried to resolve this by hierarchical encoding
\cite{serban2017hierarchical} or memory network based on attention
\cite{seo2017visual}. However, these works mostly focus on history
dialog reasoning. The work of \cite{kottur2018visual} solved the
co-reference on both visual and linguistic space using neural module
networks.   Recently, like the other vision-language tasks, yet
visual dialog is also beneficial from cross-modal pre-training which
was employed  \cite{Hong_2021_CVPR,murahari2020large,wang2020vd,Zhuge_2021_CVPR}. 

The video dialog task pushes the challenges further, thanks to the
complexity of analyzing video. Video-grounded dialog system (VGDS)
received more interested from community recently with the DSTC7 \cite{yoshino2019dialog}
and DSTC8 \cite{kim2019eighth} challenge. Early attempts \cite{hori2019end,le2019end,nguyen2018film,sanabria2019cmu}
employ the recurrent neural network to encode dialog history. Later
methods \cite{chu2020multi,lin2019entropy,yeh2019reactive} use the
Attention mechanism, or \cite{xie2020audio} design a memory networks
to extract the relationship between different modalities, \cite{le2020multimodal,lee2020dstc8,le2020bist}
employ Transformer-based network to resolve the cross modality learning.
More explicit relationships in dialogs are recently studied, showing
promising results \cite{geng2021dynamic,le2021learning}. This extends
to co-reference graph across visual and textual domains \cite{kim2021structured}.
However, these approach all represent the video by frame features,
just lacking the key concepts of of object permanence.

\paragraph{Object-centric visual reasoning}

Visual QA and visual dialog benefit greatly from object-centric representation
of visual content as this fills the gap between low-level visual features
and high-level linguistic semantics \cite{dang2021hierarchical,le2020dynamic}.
The early work proposed the object-centric representation \cite{desta2018object}
by leveraging the object detection for images and \cite{chao2018rethinking,kalogeiton2017action,wojke2017simple,xie2018rethinking,yang2020bert}
combining with the tracking algorithm for video inputs. In problems
that require further reasoning on objects, a relational network was
introduced by \cite{baradel2018object}, and \cite{huang2020location,kim2021structured,pan2020spatio,wang2018videos,yi2019clevrer,zeng2019graph}
make use of graph-based method for transparent reasoning. However,
these methods only use object's features as an additional input on
a generic network without any prior reasoning structure in space-time,
which is the key of success in further reasoning. The work in \cite{dang2021hierarchical}
introduces a generic reasoning unit through dynamically building interaction
graphs between objects in space-time derived by the question, but
this is limited to single question.

\section{Method}

\begin{figure}
\centering{}\includegraphics[width=0.9\textwidth]{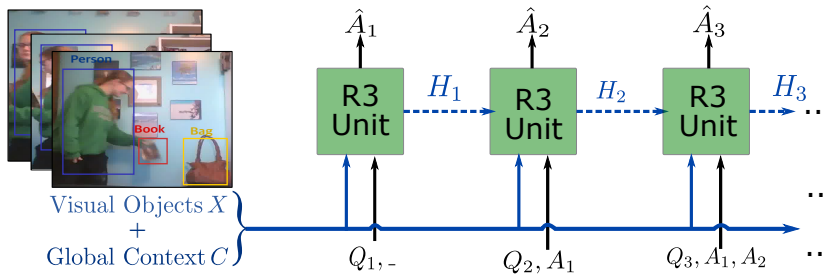}\caption{The architecture of $\protect\ModelName$ model features a chain of
\emph{Recurrent Relational Reasoning} ($\protect\ReasonUnit$) Units
which maintain the \emph{dialog states} $\left\{ H_{i}\right\} $
across the turns of the conversation. \label{fig:Model}}
\end{figure}

Given a video $V$, the task of video dialog is to hold a smooth conversation
of $T$ turns. Each turn $t$ is a textual question-answer pair $\left(Q_{t},A_{t}\right)$.
We want to estimate a model parameterized by $\theta$ that returns
the best answer for the corresponding question:
\begin{equation}
\hat{A}_{t}=\arg\max_{A}P\left(A\mid V,Q_{1:t-1},A_{1:t-1},Q_{t};\theta\right),\label{eq:dialog-model}
\end{equation}
for $t=1,2,...,T$.

The main challenge lies in the coherent answering with respect to
the existing conversation of length $t-1$, while reasoning over the
space-time of video, which itself demands efficient and effective
schemes of representation that supports a high-level dialog. With
these constraints in mind, here we treat the video $V$ as a collection
of object lives, each of which is a trajectory of spatial positions
equipped with object visual features. This object-oriented view enables
ease of constructing reasoning paths in response to linguistic queries.

In the following we present $\ModelName$, an \emph{object-centric
reasoning model} for video dialog. $\ModelName$ is a recurrent over
dialog turns, maintaining and tracking \emph{dialog states} associated
with objects. At each turn, a relational reasoning engine is called
to process the query, the video object trajectories, and the answer
history of previous turns. Fig.~\ref{fig:Model} shows the overall
architecture of $\ModelName$.

\subsection{Preliminaries}

Following \cite{dang2021hierarchical}, we parse the video of $F$
frames into $N$ sequences of objects tracked over time. Objects at
each frame are position-coded together with their appearance features.
Each \emph{visual object live} throughout the video is therefore represented
as a matrix $X_{n}\in\mathbb{R}^{F\times d}$ for $n=1,2,...,N$.
Moreover, each frame is represented by a holistic context vector $c\in\mathbb{R}^{1\times d}$,
therefore the holistic context matrix representation of video is denoted
as $C\in\mathbb{R}^{F\times d}$. Dialog sentences (questions and
answers) are split into words and then embedded into a matrix $S=\left[w_{1:L}\right]\in\mathbb{R}^{L\times d}$
where $L$ is sentence length. With a slight abuse of notation, we
use $d$ to denote the size of both visual feature vectors and linguistic
vectors for ease of reading. 

We also make use of the attention function \cite{vaswani2017attention}
defined over the triplet of \emph{query} $q\in\mathbb{R}^{1\times d}$,
\emph{keys} $K\in\mathbb{R}^{M\times d}$ and \emph{values} $V\in\mathbb{R}^{M\times d}$:
\begin{equation}
\text{Attn}\left(q,K,V\right):=\sum_{m=1}^{M}\text{softmax}_{m}\left(\frac{K_{m}W_{k}\left(qW_{q}\right)}{\sqrt{d}}^{\top}\right)V_{m}W_{v}\in\mathbb{R}^{1\times d}.\label{eq:attention}
\end{equation}
where $W_{k}$, $W_{q}$, and $W_{v}$ are learnable parameters. In
essence, this function reads out the most relevant values whose keys
match the query. Likewise, a linear projection $W\in\mathbb{R}^{d_{1}\times d_{2}}$
from $x\in\mathbb{R}^{1\times d_{1}}$ to $\mathbb{R}^{1\times d_{2}}$
is denoted as
\begin{equation}
\text{Linear}(x):=xW.\label{eq:linear-proj}
\end{equation}

\subsection{Recurrent Relational Reasoning over Space-Time and Turns}

\begin{figure}[t]
\centering{}\includegraphics[width=0.9\textwidth]{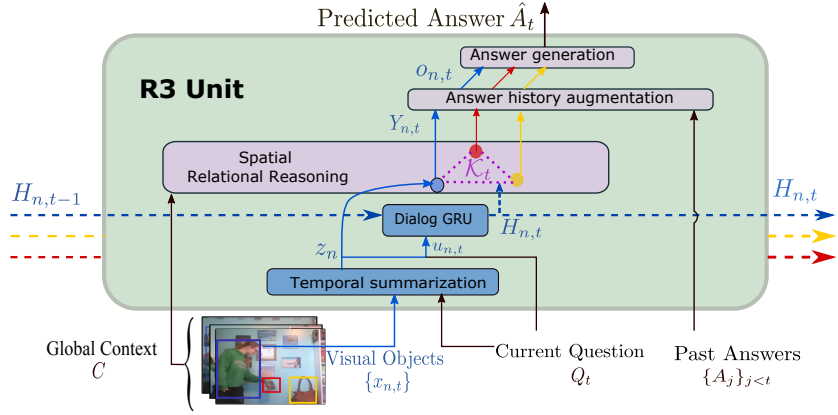}\caption{The architecture of a Recurrent Relational Reasoning ($\protect\ReasonUnit$)
Unit operating at dialog turn $t$. The space-time-dialog reasoning
happens at the three corresponding member blocks. The colors \textcolor{blue}{blue}/\textcolor{red}{red}/\textcolor{orange}{yellow}
indicate terms and operations specific to each object (only drawn
for the \textcolor{blue}{blue} object). \textcolor{magenta}{Pink}
indicates cross-object operations. \label{fig:Reasoning-unit}}
\end{figure}

The $\ModelName$ is a recurrent system that keeps track of the evolution
of the dialog states over turns. Each step is a reasoning unit $\ReasonUnit$
(short for Recurrent Relational Reasoning) which takes as input a
question $Q_{t}\in\mathbb{R}^{S\times d}$ of the present turn $t$,
the previous dialog states $H_{t-1}\in\mathbb{R}^{N\times d}$, a
holistic context representation $C\in\mathbb{R}^{F\times d}$, a set
of $N$ object sequences $X=\left\{ X_{n}\mid X_{n}\in\mathbb{R}^{F\times d}\right\} _{n=1}^{N}$,
and past answers $\left\{ A_{j}\right\} _{j<t}$. The outputs of a
$\ReasonUnit$ unit are the representations of object lives conditioned
on the current query and information from past turns. Fig.~\ref{fig:Reasoning-unit}
illustrates the structure of $\ReasonUnit$ unit. 

\subsubsection{Query-induced Temporal Summarization}

At each \emph{conversational turn }$t$, we produce query-specific
object representation. Firstly, we generate a \emph{query-specific}
summary of each object sequence $X_{n}\in\mathbb{R}^{F\times d}$
into a vector $z_{n}\in\mathbb{R}^{1\times d}$ using temporal attention
over frames. The frame attention weights are driven by the words in
the query $Q_{t}=\left\{ Q_{s,t}\in\mathbb{R}^{1\times d}\right\} _{s=1}^{S_{t}}$.
These produce a query-specific \emph{object resume} as follows:\vspace{-0.3em}

\begin{align}
z_{n} & =\frac{1}{S_{t}}\sum_{s=1}^{S_{t}}\text{Attn}\left(Q_{s,t},X_{n},X_{n}\right)\in\mathbb{R}^{1\times d}.\label{eq:obj-resume}
\end{align}
To handle the case where an object cannot be detected at particular
frames, we place a binary mask in the appropriate place.

Next we generate an \emph{object-specific} embedding of the question:
\[
q_{n}=\text{Attn}\left(z_{n},Q_{t},Q_{t}\right)\in\mathbb{R}^{1\times d},
\]
Finally the object embedding is modulated by the question as:
\begin{equation}
u_{n,t}=\text{tanh}\left(\left[z_{n},q_{n},q_{n}\odot z_{n}\right]\right)\in\mathbb{R}^{1\times3d}.\label{eq:quest-obj-embed}
\end{equation}
This embedding serves as input for the recurrent network, which is
presented in the next subsection.

\subsubsection{Recurrent Dialog States}

In dialog, the question at a turn typically advances from, and co-refers
to, the previous questions and answers. In video dialog, questions
are semantically related to objects appearing in video. We hence maintain
a \textbf{dialog state} at turn $t$ in the form of a matrix $H_{t}\in\mathbb{R}^{N\times d}$,
i.e., each row corresponds to an object. The state dynamics is modeled
in a set of $N$ parallel recurrent networks:\vspace{-0.3em}

\begin{align}
H_{n,t} & =\text{GRU}\left(H_{n,t-1},u_{n,t}\right)\in\mathbb{R}^{1\times d},\label{eq:GRU}
\end{align}
for $n=1,2,...,N$, where GRU is a standard Gated Recurrent Unit \cite{cho2014properties},
$u_{n,t}$ is turn-specific embedding of the object $n$ at turn $t$
calculated in Eq.~(\ref{eq:quest-obj-embed}). 

As the dialog states are object-centric, co-references between questions
are \emph{indirectly} and \emph{distributionally} captured into the
current multi-object states through the integration of previous states
$H_{n,t-1}$ (which contains information of the previous questions)
and the current question as part of $u_{n,t}$. In what follows, we
will show how $H_{t}$ is used for relational reasoning.

\subsubsection{Relational Reasoning between Objects}

Equipped with dialog states, we now model the inter-object interaction
which describes the behavior of objects with their neighbors driven
by the current question $Q_{t}$. We employ a spatial graph whose
vertices are the object resumes $Z=\left\{ z_{n}\right\} _{n=1}^{N}$
computed in Eq.~(\ref{eq:obj-resume}), and the edges are represented
by an adjacency matrix $\mathcal{K}_{t}\in\mathbb{R}^{N\times N}$
which is calculated dynamically as the \emph{question-specific interaction
matrix }between objects:
\begin{equation}
\mathcal{K}_{t}=\text{softmax}\left(\frac{H_{t}H_{t}^{\top}}{\sqrt{d}}\right),\label{eq:adjacency-matrix}
\end{equation}
where $H_{t}$ is computed in Eq.~(\ref{eq:GRU}). This matrix serves
as a backbone for a Deep Graph Convolutional Network (DGCN) \cite{le2020dynamic}
to refine object representations by taking into account the relations
with their neighboring nodes:
\begin{equation}
\bar{Z}=\text{DGCN}\left(Z;\mathcal{K}_{t}\right)\label{eq:DCGN}
\end{equation}
Details of $\text{DCGN}\left(\cdot;\cdot\right)$ is given in the
Supplement.

\paragraph{Utilizing visual context }

To utilize the underlying background scene information and compensate
for possible undetected objects, we augment the object representations
with the holistic context information $C\in\mathbb{R}^{F\times d}$.
We summarize the context sequence into a vector as follows:

\begin{equation}
\bar{c}=\frac{1}{S_{t}}\sum_{s=1}^{S_{t}}\text{Attn}\left(Q_{s,t},C,C\right)\in\mathbb{R}^{1\times d}.\label{eq:R3-global-context}
\end{equation}
Finally, the final output of this component is

\begin{equation}
Y_{n,t}=\text{Linear}\left(\left[\bar{Z}[n];\bar{c}\right]\right)\in\mathbb{R}^{1\times d},\label{eq:R3-output}
\end{equation}
where $\bar{Z}[n]$ is the $n$-th row of $\bar{Z}$ in Eq.~(\ref{eq:DCGN}).

\subsubsection{Leveraging Answer History}

So far, the $\ReasonUnit$ unit has used the recurrent dialog states
to generate the representation of objects for the current question,
however, part of historical information has been forgotten after a
long conversation. It is therefore necessary to maintain a dynamic
history of previous turns, which will be queried when seeking an answer
to a new question. This would help to mitigate the co-reference effect
by borrowing parts of previous answers into the current answer when
deemed relevant.

Recall that the reasoning by the $\ReasonUnit$ unit results in each
object having a turn-specific representation $Y_{n,j}\in\mathbb{R}^{1\times d}$
for turn $j=1,2,...,t-1$. Let $A_{j-1}\in\mathbb{R}^{L_{j-1}\times d}$
be the embedding the answer at turn $j-1$. The \emph{object-specific
answer embedding} is computed as:\vspace{-0.4em}

\begin{align}
a_{n,j-1} & =\text{Attn}\left(Y_{n,j},A_{j-1},A_{j-1}\right)\in\mathbb{R}^{1\times d}.
\end{align}
This is combined with the object representation and turn position
to generate a new object representation:\vspace{-0.4em}

\begin{align}
G_{n,j} & =\text{Linear}\left(\left[Y_{n,j},a_{j-1},p_{j}\right]\right)\in\mathbb{R}^{1\times d},
\end{align}
where $p_{j}$ is a positional encoding feature for each turn.

Thus the collection over turns $\mathfrak{M}_{n,t}=\left[G_{n,j}\right]_{j=1}^{t-1}\in\mathbb{R}^{(j-1)\times d}$
is a history of answer-guided representation for object $n$ over
previous turns. The answer history enables the retrieval of relevant
pieces w.r.t. the current question at turn $t$:

\begin{equation}
\mathcal{H}_{n,t}=\text{Attn}\left(Y_{n,t},\mathfrak{M}_{n,t},\mathfrak{M}_{n,t}\right).\label{eq:past-reasoning}
\end{equation}
This is then augmented with the current object representation $Y_{n,t}$
in Eq.~(\ref{eq:R3-output}) to produce the final form:
\begin{equation}
O_{n,t}=\text{Linear}\left(\left[\mathcal{H}_{n,t},Y_{n,t}\right]\right);\quad\quad O_{t}=\left[O_{n,t}\right]_{n=1}^{N}\in R^{N\times d}.\label{eq:COST-output}
\end{equation}
This serves as input for the answer generation module, which we present
next.

\subsection{Answer Generation}

\begin{figure}[t]
\centering{}\includegraphics[width=1\textwidth]{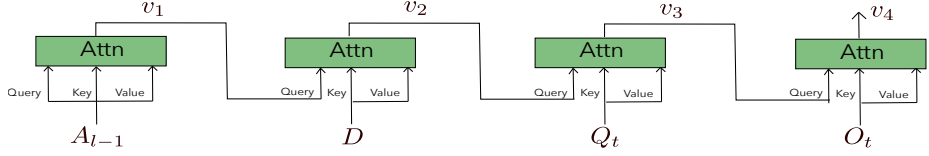}\caption{Four-step Transformer decoder used in the Answer generator for the
question at turn $t$ at generation step $l$. $A_{l-1}$: embedding
matrix of the unfinished utterance of length $l-1$; $D$: embedding
matrix of the dialog history; $Q_{t}:$ embedding of the question
at turn $t$; $O_{t}$: output of $\protect\ModelName$.\label{fig:Answer-generator}}
\end{figure}

To generate the response utterance, we employ the standard autoregressive
framework to produce one word at a time by iteratively estimating
the conditional word distribution $P\left(w\mid w_{t,1:l-1};V,Q_{1:t-1},A_{1:t-1},Q_{t}\right)$
at generation step $l$. Inspired by the decoders in \cite{le2019multimodal,le2020bist},
we use a four-step Transformer decoder, as illustrated in Fig.~\ref{fig:Answer-generator}.

Let $A_{t,l}\in\mathbb{R}^{l\times d}$ be the embedding matrix of
the unfinished utterance of length $l$; $D=\left[Q_{1:t-1},A_{1:t-1}\right]\in\mathbb{R}^{L_{D}\times d}$
the embedding of the dialog history of length $L_{D}$ (words); $Q_{t}\in\mathbb{R}^{L_{Q}\times d}$
the embedding of the question, and $O_{t}\in\mathbb{R}^{N\times d}$
the output of generated by the $\ModelName$ in Eq.~(\ref{eq:COST-output}).
The decoder generates a representation $v_{4}$ through a step-wise
manner:\vspace{-0.3em}

\begin{align}
v_{1} & =\text{Attn}\left(a_{l},A_{t,l-1},A_{t,l-1}\right);\qquad v_{2}=\text{Attn}\left(v_{1},D,D\right);\nonumber \\
v_{3} & =\text{Attn}\left(v_{2},Q_{t},Q_{t}\right);\qquad\hspace{1em}\hspace{1em}\,\,\,\,\,v_{4}=\text{Attn}\left(v_{3},O_{t},O_{t}\right).\label{eq:answer_generator}
\end{align}
where $a_{l}=A_{t,l}\left[l\right]$, e.g., the last row of $A_{t,l}$.
In essence, the sequence of current utterance, the existing dialog,
and the current question forms the context to query the object representations
$O_{t}$.

The retrieved information $v_{4}$ is used to generate the next word
through the word distributions:\vspace{-0.5em}

\begin{align}
P_{vocab} & =\text{softmax}\left(\text{Linear}(v_{4})\right)\in\mathbb{R}^{1\times N_{vocab}},\label{eq:P-vocab}\\
P_{q} & =\text{Ptr}\left(Q_{t},v_{4}\right)\in\mathbb{R}^{1\times N_{vocab}},\label{eq:pointer}\\
P_{l} & =\alpha P_{q}+\left(1-\alpha\right)P_{vocab},\label{eq:pointer combine}
\end{align}
where $\alpha\in(0,1)$, and $\text{Ptr}$ is a trainable Pointer
Network \cite{vinyals2015pointer} that ``points'' to all the tokens
in question $Q_{t}$ that are related to $v_{4}$, and $P_{l}$ is
a short-form for $P\left(w\mid w_{t,1:l-1},V,Q_{1:t-1},A_{1:t-1},Q_{t};\theta\right)$.
$\text{Ptr}$ seeks to reuse the relevant question words for the answer;
and this is useful when facing rare words or word repetition is required.
The gating function $\alpha$ is learnable (detailed in the Supplement).

\subsection{Training}

Given ground-truth answers $A_{1:T}^{*}$ of a full conversation of
$T$ turns, where $A_{t}^{*}=\left(w_{1},w_{2},...,w_{L_{t}^{a}}\right)$.
To make our generator more robust, we also add the log-likelihood
of re-generated current turn's question $Q_{t}^{*}=\left(w_{1},w_{2},...,w_{L_{t}^{q}}\right)$
to our loss. The network is trained with the cross-entropy loss w.r.t
parameter $\theta,\theta_{q}$ :\vspace{-0.2em}

\begin{align}
\mathcal{L} & =\mathcal{L}\left(\theta\right)+\mathcal{L}\left(\theta_{q}\right),\text{where}\\
\mathcal{L}\left(\theta\right) & =\sum_{t=1}^{T}\text{log}P\left(A_{t}^{*}\mid V,Q_{1:t-1},A_{1:t-1}^{*},Q_{t};\theta\right)\\
 & =\sum_{t=1}^{T}\sum_{l=1}^{L_{t}^{a}}\text{log}P_{l}\left(w_{l}\mid w_{t,1:l-1},V,Q_{1:t-1},A_{1:t-1}^{*},Q_{t};\theta\right);\text{and}\\
\mathcal{L}\left(\theta_{q}\right) & =\text{log}P_{vocab}\left(Q_{t}^{*}\mid V,Q_{t};\theta_{q}\right)\\
 & =\sum_{l=1}^{L_{t}^{q}}\text{log}P_{vocab}\left(w_{l}^{q}\mid w_{t,1:l-1}^{q},V,Q_{t};\theta_{q}\right);
\end{align}
where $P_{l}\left(\cdot\right),P_{vocab}\left(\cdot\right)$ are computed
in Eqs.~(\ref{eq:P-vocab}--\ref{eq:pointer combine}).

\section{Experiments}

\subsection{Experimental Settings}

\paragraph{Dataset:}

We train our proposed method $\ModelName$ on the Audio-Visual Scene-Aware
Dialogue (AVSD) \cite{alamri2019audio}, a large-scale video grounded
dialog dataset. The dataset provides text-based dialogs visually grounded
on untrimmed action videos from the popular Charades dataset \cite{sigurdsson2016hollywood}.
Each annotated dialog consists of 10 rounds of question-answer about
both static and dynamic scenes in a video, including objects, actions,
and audio content. We benchmark against existing methods on video
dialog using two different test splits used at the Seventh Dialog
System Technology Challenges (DSTC7) \cite{yoshino2019dialog} and
the Eighth Dialog System Technology Challenges (DSTC8) \cite{kim2019eighth}.
See Table~\ref{tab:Statistics-of-AVSD} for detailed statistics of
the AVSD dataset and the two test splits at DSTC7 and DSTC8.

Often state-of-the-art methods in video dialog rely on different sources
of information, including visual dynamic scenes, text description
in the form of caption/video summary, and audio content. While the
textual data containing high-level information of visual content can
significantly attribute to the model's performance, they provide shortcuts
due to linguistic biases that the models can exploit. As the ultimate
goal of the video dialog task is to benchmark if a model can gather
the visual clues to produce an appropriate response for a smooth conversation
with humans, our experiments deliberately aim at challenging the visual
reasoning capability of the models. In particular, we assume that
the models only have access to the visual content and dialog history
to answer a question at a specific turn while ignoring other additional
textual data and audio data used by many other methods \cite{le2019multimodal,le2020bist,nguyen2018film}.

\begin{table}[t]
\centering{}\caption{Statistics of the AVSD dataset and the test splits used at DSTC7 and
DSTC8.\label{tab:Statistics-of-AVSD}}
\begin{tabular}{l|cccc}
\cline{2-5} \cline{3-5} \cline{4-5} \cline{5-5} 
 & Training & Validation & DSTC7 Test & DSTC8 Test\tabularnewline
\hline 
\hline 
No. Videos & 7,659 & 1,787 & 1,710 & 1,710\tabularnewline
No. Dialog turns & 153,180 & 35,740 & 13,490 & 18,810\tabularnewline
\hline 
\end{tabular}
\end{table}

\paragraph{Implementation details:}

We implement our models with PyTorch. All models are trained by optimizing
the multi-label cross-entropy loss over generated tokens using Adam
optimizer \cite{kingma2014adam} with cosine learning rate scheduler
\cite{loshchilov2016sgdr}. At the training stage, we also use the
auto-encoder loss function for the current question from \cite{le2019multimodal}.
We use a batch size of 128 samples distributed on 4 GPUs and train
all the models for 50 epochs. Unless stated otherwise, each attention
component in Eq.~(\ref{eq:answer_generator}) is composed of a stack
of 3 identical attention layers in Eq.~(\ref{eq:attention}). For
each attention layer, we also use 4 parallel heads as suggested by
\cite{vaswani2017attention}. Model parameters are selected based
on the convergence of validation loss. At inference time, we adopt
a beam search algorithm with a beam size of 3 for our answer generator.

Regarding object lives extraction, we strictly follow \cite{dang2021hierarchical}
and extract 30 object sequences per video. We further apply frame
sub-sampling at an overall ratio of 4:1 to reduce the computational
expensiveness. On average, each object live is composed of 176 time
steps. For the context features $C$ used in Eq.~(\ref{eq:R3-global-context}),
we use I3D features \cite{carreira2017quo} similar to other existing
methods \cite{le2020bist,le2020multimodal}. Pytorch implementation
of our model is available online\footnote{https://github.com/hoanganhpham1006/COST}.

\paragraph{Evaluation metrics:}

We adopt the same word-overlap-based metrics, including BLEU, METEOR,
ROUGE-L, and CIDEr, as used by \cite{yoshino2019dialog} to evaluate
the effectiveness of the models. Results of prior methods are reported
in the respective papers or by using the official source code.

\subsection{Comparison against SOTAs\label{subsec:Comparison-against-SOTAs}}

We compare against the state-of-the-art methods, including MTN \cite{le2019multimodal},
FA+HRED \cite{nguyen2018film}, Student-Teacher \cite{hori2019joint},
SCGA \cite{kim2021structured} and BiST \cite{le2020bist} on both
AVSD@DSTC7 and AVSD@DSTC8 test splits. For fair comparisons, all models
only make use of the video content and dialog history. Results are
shown in Table~\ref{tab:Results_DSTC7} and Table~\ref{tab:Results_DSTC8}
for DSTC7 and DSTC8 test splits, respectively. In particular, $\ModelName$
consistently sets new SOTA performance on all evaluation metrics against
existing methods on both test splits. The results strongly demonstrate
the efficiency of our object-centric reasoning model with recurrent
relational reasoning compared to methods that rely only on holistic
visual features such as I3D \cite{carreira2017quo} and ResNeXt \cite{xie2017aggregated}. 

\begin{table}
\centering{}\caption{Experimental results on the AVSD@DSTC7 test split. All models only
have access to video content and dialog history. $^{\dagger}$Models
use visual features other than holistic video features such as I3D
or ResNeXt. $\protect\ModelName$ use object sequences and I3D features.\label{tab:Results_DSTC7}}
\begin{tabular}{l|ccccccc}
\hline 
Methods & BLEU1 & BLEU2 & BLEU3 & BLEU4 & METEOR & ROUGE-L & CIDEr\tabularnewline
\hline 
\hline 
FA+HRED \cite{nguyen2018film} & 0.648 & 0.505 & 0.399 & 0.323 & 0.231 & 0.510 & 0.843\tabularnewline
MTN (I3D) \cite{le2019multimodal} & 0.654 & 0.521 & 0.420 & 0.343 & 0.247 & 0.520 & 0.936\tabularnewline
MTN (ResNeXt) \cite{le2019multimodal} & 0.688 & 0.55 & 0.444 & 0.363 & 0.260 & 0.541 & 0.985\tabularnewline
Student-Teacher \cite{hori2019joint} & 0.675 & 0.543 & 0.446 & 0.371 & 0.248 & 0.527 & 0.966\tabularnewline
BiST \cite{le2020bist} & 0.711 & 0.578 & 0.475 & 0.394 & 0.261 & 0.550 & 1.050\tabularnewline
SCGA$^{\dagger}$ \cite{kim2021structured} & 0.702 & 0.588 & 0.481 & 0.398 & 0.256 & 0.546 & 1.059\tabularnewline
\hline 
$\ModelName$ (Ours) & \textbf{0.723} & \textbf{0.589} & \textbf{0.483} & \textbf{0.400} & \textbf{0.266} & \textbf{0.561} & \textbf{1.085}\tabularnewline
\hline 
\end{tabular}\vspace{-0.2em}
\end{table}

\begin{table}
\centering{}\caption{Experimental results on the AVSD@DSTC8 test split.\label{tab:Results_DSTC8}}
\begin{tabular}{l|ccccccc}
\hline 
Methods & BLEU1 & BLEU2 & BLEU3 & BLEU4 & METEOR & ROUGE-L & CIDEr\tabularnewline
\hline 
\hline 
MTN (I3D) \cite{le2019multimodal} & 0.611 & 0.496 & 0.404 & 0.336 & 0.233 & 0.505 & 0.867\tabularnewline
MTN (ResNeXt) \cite{le2019multimodal} & 0.643 & 0.523 & 0.427 & 0.356 & 0.245 & 0.525 & 0.912\tabularnewline
BiST \cite{le2020bist} & 0.684 & 0.548 & 0.457 & 0.376 & 0.273 & 0.563 & 1.017\tabularnewline
SCGA$^{\dagger}$ \cite{kim2021structured} & 0.675 & 0.559 & 0.459 & 0.377 & 0.269 & 0.555 & 1.024\tabularnewline
\hline 
$\ModelName$ (Ours) & \textbf{0.695} & \textbf{0.559} & \textbf{0.465} & \textbf{0.382} & \textbf{0.278} & \textbf{0.574} & \textbf{1.051}\tabularnewline
\hline 
\end{tabular}\vspace{-0.2em}
\end{table}

\subsection{Model Analysis\label{subsec:Model-analysis}}

\subsubsection{Object-centric Representation Facilitates Space-Time-Dialog Reasoning}

\begin{figure}[t]
\begin{centering}
\noindent\begin{minipage}[t]{1\columnwidth}%
\begin{center}
\begin{minipage}[t]{0.45\columnwidth}%
\begin{center}
\includegraphics[width=0.95\columnwidth,height=0.9\columnwidth]{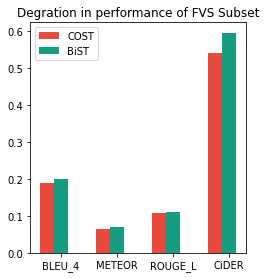}\\
(a)
\par\end{center}%
\end{minipage}\quad{}%
\begin{minipage}[t]{0.45\columnwidth}%
\begin{center}
\includegraphics[width=0.95\columnwidth,height=0.9\columnwidth]{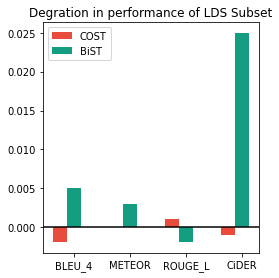}\\
(b)
\par\end{center}%
\end{minipage}
\par\end{center}%
\end{minipage}
\par\end{centering}
\caption{Performance degradation points from DSTC7 full test splits to (a)
FVS subset and (b) LDS subset. The FVS requires fine-grained visual
understanding to answer questions while the LDS challenges the capability
to handle long-distance dependencies questions of the models. The
lower the better. Negative degradation points indicate improvements
in performance. $\protect\ModelName$ demonstrates its robustness
against degradation in performance compared to BiST on these challenging
subsets.\label{fig:Performance-degradation} }

\end{figure}

To better highlight the effectiveness of our object-centric representation
for reasoning over space-time to explore the semantic structure in
videos, we design a subset of the AVSD@DSTC7 test split that poses
challenges for models heavily relying on linguistic biases but undervaluing
fine-grained visual information. First, we train a variant of the
MTN model \cite{le2019multimodal} with all the visual and audio components
removed on the AVSD dataset. Next, we only pick dialog turns and
their associated videos having BLEU4 score lower than 0.05 on the
DSTC7 test split. This eliminates any questions whose answers can
be guessed by the linguistic biases. Eventually, we obtain a subset
of 1,062 videos with 8,782 associated dialog turns referred to as\emph{
Fine-grained Visual Subset (FVS)}. We evaluate the degradation of
$\ModelName$ and the current state-of-the-art BiST on the FVS subset
and report the results in Fig.~\ref{fig:Performance-degradation}(a).
As shown, $\ModelName$ is more robust to questions in FVS while BiST
struggles as it is degraded by larger margins on all the evaluation
metrics. The results are clearly evident the benefits of the fine-grained
video understanding brought by the object-centric representation compared
to the holistic video representation used by MTN and BiST.

\subsubsection{Recurrent Modeling Supports Long-Distance Dependencies}

One of the main advantages of $\ModelName$ against SOTA methods is
that it maintains a \emph{recurrent system of} \emph{dialog states,}
which offers the better capability of handling long-distance dependencies
questions. These questions require models to maintain and retrieve
information appearing far in earlier turns. Methods without an explicit
mechanism to propagate long-distance dependencies would struggle to
generalize. In order to verify this, we design another subset of the
AVSD@DSTC7 test split where we only collect questions at turns greater
than 3. This results in a subset of 950 videos and 11,210 associated
dialog turns. We refer to this as \emph{Long-distance Dependencies
Subset (LDS)}. Fig.~\ref{fig:Performance-degradation}(b) details
the degradation in performance of $\ModelName$ and BiST on the LDS.
As shown, while $\ModelName$ achieves slight improvements in performance
(negative degradation) thanks to it recurrent design, BiST experiences
consistent losses across the evaluation metrics. The results verify
the validity of our hypothesis that maintaining the recurrent dialog
object states is beneficial for handling long-distance dependencies
between turns.

\subsection{Ablation Studies}

\begin{table}[t]
\centering{}\caption{Ablation studies on AVSD@DSTC7 test split. AHR: Answer history retrieval.\label{tab:Ablation}}
\begin{tabular}{l|ccccccc}
\hline 
Effects of & BLEU1 & BLEU2 & BLEU3 & BLEU4 & METEOR & ROUGE-L & CIDEr\tabularnewline
\hline 
\hline 
\textbf{recurrent design} &  &  &  &  &  &  & \tabularnewline
\hspace{2em}w/o recurrence & 0.710 & 0.577 & 0.471 & 0.388 & 0.261 & 0.554 & 1.041\tabularnewline
\textbf{object modeling} &  &  &  &  &  &  & \tabularnewline
\hspace{2em}w/o object-centric & 0.709 & 0.574 & 0.467 & 0.385 & 0.260 & 0.553 & 1.042\tabularnewline
\textbf{attention in AHR} &  &  &  &  &  &  & \tabularnewline
\hspace{2em}w/o self-attention & 0.719 & 0.584 & 0.477 & 0.395 & 0.263 & 0.558 & 1.062\tabularnewline
\textbf{pointer networks} &  &  &  &  &  &  & \tabularnewline
\hspace{2em}w/o pointer & 0.715 & 0.583 & 0.477 & 0.394 & 0.260 & 0.557 & 1.045\tabularnewline
\hline 
\textbf{Full model} & \textbf{0.723} & \textbf{0.589} & \textbf{0.483} & \textbf{0.400} & \textbf{0.266} & \textbf{0.561} & \textbf{1.085}\tabularnewline
\hline 
\end{tabular}
\end{table}

We conduct an extensive set of ablation studies to examine the contributions
of each components in $\ModelName$ (See Table~\ref{tab:Ablation}).
These include ablating the recurrent design of $\ModelName$, the
use of object-centric video representation, and the use of self-attention
layers for answer history retrieval for answer generation. We find
that ablating any of these components would take the edge off the
model's performance. The results are consistent with our analysis
in Sec.~\ref{subsec:Model-analysis} on the crucial effects of our
recurrent design and the object-centric modeling by $\ModelName$
on the overall performance. We detail the effects as follows.

\emph{Effect of recurrent design}: We remove the GRU in Eq.~(\ref{eq:GRU})
and use query-specific object representation outputs of Eq.~(\ref{eq:quest-obj-embed})
as direct input to compute the adjacency matrix $\mathcal{K}_{t}$
in Eq.~(\ref{eq:adjacency-matrix}). By doing this, we ignore the
effect of past dialog states on the current turn. As clearly seen,
the model's performance considerably degrades on all evaluation metrics.

\emph{Effect of object-centric modeling}: In this experiment, we use
only the context feature $C$ (I3D features) and remove all the effects
of the object representation. This leads to performance degradation
by nearly 2\% on all BLEU scores. The fine-grained object-centric
representation clearly has an effect on improving the understanding
of hidden semantic structure in video.

\emph{Effect of self-attention-based answer history retrieval}: This
experiment ablates the role of prior generated answer tokens on information
retrieval as in Eq.~(\ref{eq:past-reasoning}). We instead use the
turn-specific visual representation as a direct input for the answer
generator. The results show that this slightly affects the overall
performance of the model.

\emph{Effect of pointer networks for answer generation}: We remove
the use of the pointer networks as in Eqs.~(\ref{eq:pointer} and
\ref{eq:pointer combine}) during answer generation. The results show
that the removal of the pointer slightly hurt the model's performance.

\subsection{Qualitative Analysis}

\begin{figure}[t]
\centering{}\includegraphics[width=0.9\textwidth]{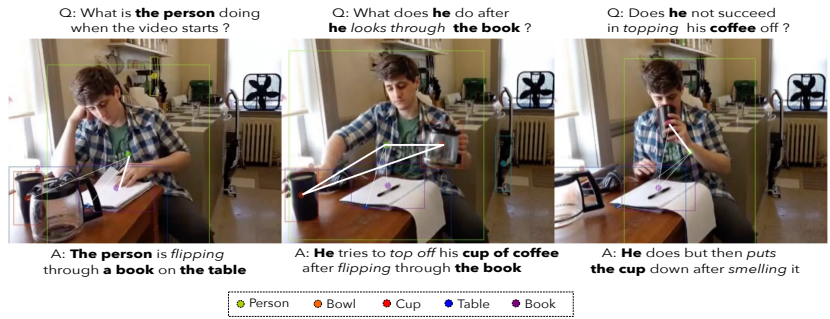}\caption{Visualization of the question-specific interaction matrices between
objects $\mathcal{K}_{t}$ of Eq.~(\ref{eq:adjacency-matrix}). Each
frame/question pair (left to right) represents a dialog turn, where
the frame is selected to reflect the moment being queried. The visibility
of edges denotes the relevance of visual objects' relations to the
questions and desired answers. Detected objects are named by Faster
RCNN. $\protect\ModelName$ succeeds in constructing turn-specific
graphs of relevant visual objects that facilitate answering questions.
Sample is taken from DSTC7 test split - Best viewed in colors.\label{fig:visualization}}
\end{figure}

We visualize a representative example from the AVSD@DSTC7 test split
as a showcase to analyze the internal operation of the proposed method
$\ModelName$. We present the question-induced interaction matrix
as it is a crucial component of our model design in Eq.~(\ref{eq:adjacency-matrix}).
Fig.~\ref{fig:visualization} presents the evolution of the relationships
between objects in video from turn to turn (left to right). $\ModelName$
succeeds in constructing turn-specific graphs of relevant visual objects
that reflect the relationships of interest by respective questions
and answers. The interpretability and strong qualitative results (Sec.~\ref{subsec:Comparison-against-SOTAs}
and \ref{subsec:Model-analysis}) by $\ModelName$ are evident to
the appropriateness of the object-centric representation towards solving
video dialog task.

\section{Conclusion}

Addressing the highly challenging task of video dialog, we have proposed
$\ModelName$, a new \emph{recurrent object-centric} system that learns
to reason through multiple dialog turns, object dynamics, and interactions
over space-time in video. $\ModelName$ maintains and tracks dialog
states over the course of conversation. It treats objects in video
as primitive constructs whose ``lives'' and relations to others
throughout the video are dynamically examined through the guidance
of the questions, conditioned on the \emph{dialog states} and \emph{answer
history}. Object representation is iteratively refined at each turn,
taking into account other objects in the same spatio-temporal context,
the dialog states, the current questions and previous answers. Co-references
between concepts across dialog turns are thus handled implicitly through
dynamic concept-object binding. Tested on the challenging AVSD dataset,
$\ModelName$ demonstrates its effectiveness against state-of-the-art
models.  Future work will explore new ways to incorporate \emph{intrinsic}
relational reasoning in the recurrent networks of dialog states and
thus address the co-reference more directly, and with more sophisticated
contextual, temporal features.

\bibliographystyle{splncs04}
\bibliography{video-dialog}

\end{document}